\documentclass[runningheads]{llncs}
\usepackage{graphicx}
\usepackage{hyperref}
\usepackage{amssymb}
\usepackage{amsmath}
\usepackage{multirow}
\usepackage{wrapfig}
\usepackage{todonotes}
\usepackage{booktabs}

\newcommand{\pt}{\textsc{ProcessTransformer} }

\begin{document}
\title{~\textsc{ProcessTransformer}:  Predictive Business Process Monitoring with Transformer Network}
\author{Zaharah A. Bukhsh\href{https://orcid.org/0000-0003-3037-8998}{\includegraphics[scale=1]{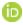}}
\and
Aaqib Saeed\href{https://orcid.org/0000-0003-1473-0322}{\includegraphics[scale=1]{Figures/orcid.JPG}} \and Remco M. Dijkman\href{https://orcid.org/0000-0003-4083-0036}{\includegraphics[scale=1]{Figures/orcid.JPG}}}
\authorrunning{Z. A. Bukhsh et al.}
\titlerunning{Predictive Business Process Monitoring with Transformer Network}
\institute{
Eindhoven University of Technology, Eindhoven, The Netherlands \\
\email{\{z.bukhsh, a.saeed, r.m.dijkman\}@tue.nl}}

% \institute{\text{\textwidth School of Industrial Engineering, Eindhoven University of Technology, Eindhoven, Netherlands}\\ 
%  \and
% \text{\textwidth Department of Mathematics and Computer Science, Eindhoven University of Technology}\\
% \email{\{z.bukhsh, a.saeed\}@tue.nl}}
%
\maketitle              % typeset the header of the contribution
\begin{abstract}
Predictive business process monitoring focuses on predicting future characteristics of a running process using event logs.
The foresight into process execution promises great potentials for efficient operations, better resource management, and effective customer services. 
Deep learning-based approaches have been widely adopted in process mining to address the limitations of classical algorithms for solving multiple problems, especially the next event and remaining-time prediction tasks.
Nevertheless, designing a deep neural architecture that performs competitively across various tasks is challenging as existing methods fail to capture long-range dependencies in the input sequences and perform poorly for lengthy process traces.
In this paper, we propose~\textsc{ProcessTransformer}, an approach for learning high-level representations from event logs with an attention-based network. 
Our model incorporates long-range memory and relies on a self-attention mechanism to establish dependencies between a multitude of event sequences and corresponding outputs. 
We evaluate the applicability of our technique on nine real event logs.
We demonstrate that the transformer-based model outperforms several baselines of prior techniques by obtaining on average above 80\% accuracy for the task of predicting the next activity. Our method also perform competitively, compared to baselines, for the tasks of predicting event time and remaining time of a running case.  % with a significant margin. 

%Finally, we show that the learned representations with the~\textsc{ProcessTransformer} can be effectively leveraged for a diverse set of process improvement tasks.

\end{abstract}
\section{Introduction} 
With the trend towards digital transformation and the availability of relatively cheaper storage solutions, the amount of process execution data (also referred to as event logs) are increasing at a tremendous scale. 
Process mining methods have been used to discover, monitor, and improve the business processes by analyzing the process logs.%~\cite{van2016data}. 
~Instead of post-hoc analysis, organizations are actively investing in predictive analytics solutions to gain insights into their performance. 
Predictive business process monitoring (PBPM) has emerged as a crucial area of process mining, focusing on estimating future characteristics of a running business process. 
PBPM has several useful business applications, including effective resource management, improving operational efficiency, and avoiding deadlock by \textit{predicting the next possible activities, duration, and remaining time to completion}. 

%Classical process analytic techniques (e.g. ~\cite{van2008cycle,van2011time,rogge2013prediction}) rely on models of formal languages such as (stochastic) Petri-nets and non-parametric methods to capture the process execution. The formal methods based mining techniques have performed effectively for process discovery and modeling, but they are limited in their capabilities to support predictive process monitoring tasks. The key limitations of traditional process analytic techniques include challenges of scalability and inefficiency for managing large-scale event data, the inapplicability of spaghetti-like process models~\cite{van2011process} for the scope of process monitoring as also noted by~\cite{pasquadibisceglie2019using}, and the limitation of generalizability for various prediction tasks.  

%deep learning methods have achieved state-of-the-art performance on challenging problems, namely object recognition, speech detection, natural language processing, and sound understanding. Likewise,

In the last decade, deep neural networks are widely adopted for tasks related to business process monitoring. 
Evermann et al.,\cite{evermann2016deep} introduced recurrent neural networks (RNNs) for predicting the process behavior at run-time. 
Following this, several prediction models based on RNNs and related variants, such as long-short term memory (LSTM) networks have been proposed for the tasks of next activity~\cite{tax2017predictive,nguyen2020time,taymouri2020predictive}, suffix generation~\cite{lin2019mm,camargo2019learning}, outcome prediction~\cite{teinemaa2019outcome}, and process's remaining time prediction~\cite{polato2018time,navarin2017lstm}. %Besides efficiency and scalability, the deep models improve predictive performance by automatically extracting underlying useful features directly from the raw data.  
Even though deep RNNs based PBPM methods have been firmly established for event sequence modeling, they suffer from considerable shortcomings, specifically for the next activity prediction task. 

Firstly, several proposed methods~\cite{tax2017predictive,camargo2019learning,navarin2017lstm,nguyen2020time,taymouri2020predictive} employ one-hot encoding to obtain the numeric representations of categorical events sequences. 
These integer representations disregard the intrinsic relationship among events and introduce unrealistic computational requirements due to an increase of data dimensionality~\cite{guo2016entity}. 
Secondly, LSTM lacks the explicit modeling of long and short-range dependencies in the sense that their performance degrades in proportion to the length of events sequences~\cite{paperno2016lambada}. 
It is specifically undesired for event logs due to interconnections introduced by control flows among activities. 
Lastly, the inherent sequential nature of LSTM and RNNs precludes parallelization, resulting in critically inefficient learning and inference. 

The attention mechanism is proposed to address the problem of long-range dependencies for sequence modeling without regard to their distance in input and output sequence~\cite{bahdanau2014neural}.
In particular, Vaswani et al.~\cite{vaswani2017attention} introduced Transformer neural network architecture, a deep sequence model that employs self-attention to maintain coherence in long-range sequences. 
The Transformer-based encoder-decoder models have rapidly become a dominant architecture for neural machine translation and natural language understanding~\cite{wolf2020transformers}. 
Specifically, Transformer architecture is also behind the compelling language models, such as GPT-3 (Generative Pretrained Transformer) and BERT (Bidirectional Encoder Representations from Transformers) that have revolutionized numerous language understanding tasks.
An interested reader may refer to~\cite{bahdanau2014neural} for a comprehensive explanation about attention-based networks.  %We provide details of Transformer architecture and self-attention mechanism in the context of PBPM in Section~\ref{sec:approach}.   

% \begin{figure}[!b]
% \centering
% \includegraphics[scale = 2,width=\textwidth]{Figures/overview_pt.pdf}
% \caption{Predictive business process monitoring with Process Attention Transformer.}
% \label{fig:overview}
% \end{figure}

Despite showing remarkable performance in multiple sequence modeling problems, Transformers have not been explored in the realm of business process management.
Thus, the core contribution of our work is the~\pt model with an improved strategy for learning high-level generic representations directly from temporal sequential events with minimal preprocessing of the input. 
% Figure~\ref{fig:overview} provides a schematic view of our approach. 
% It highlights that the business information system generates event logs based on a process model. 
% The~\pt takes the event logs as input (and other temporal features depending on the task) to predict the process monitoring tasks' outputs, including next activity, its time, and remaining time to completion. 
In contrast to fixed-size recurrent memory models, the self-attention mechanism enables access to any part of the previously generated events in a sequence. 
It allows the deep neural network to capture global dependencies between inputs and outputs for powerful general-purpose representation learning. 
Besides,~\pt can effectively differentiate the most relevant features that affect the model prediction. 
We demonstrate the applicability of~\pt on nine real event logs. We show that our Transformer-based architecture outperforms several baselines of existing methods on predicting the next activity task with minimal data preprocessing. 
Similarly,~\pt also shows performance improvements for predicting the next event time and completion time of a running case.

The remainder of the paper is structured as follows. Section~\ref{sec:rw} provides the background and related work. Section~\ref{sec:pre} provides definitions of main concepts related to PBPM. Section~\ref{sec:approach} presents the proposed approach, followed by Section~\ref{sec:evaluation}, which details the experimental setup and key results. Finally, Section~\ref{sec:conclusion} summarizes the findings, contributions and outlines the future work.   

%\section{Background and Related work}

% \subsection{Sequence modeling and neural networks}
% - PARA 1 - What is sequence modeling... how neural models can help
% - PARA 2 - What is RNN AND HOW IT IS NOT ENOUGHT
% -PARA 3 - What is LSTM and how it is not enoght
% -PARA 4 - What is attention and how it can help. 
% VIDEO ON YOUTUBE

\section{Background and Related work}
\label{sec:rw}
Representation learning focuses on extracting discriminative features from raw unstructured data to effectively solve the prediction problem, ideally in an end-to-end manner. The general-purpose representations learned from a plethora of raw, real-life event logs can be used to solve several business tasks of interest, including predictive process monitoring, improvement, and enhancement. 

In this section, we introduce sequence modeling in the context of deep learning. We also provide a brief overview of literature studies related to PBPM. 
\subsection{Sequence modeling using deep neural architectures}
 Sequence modeling involves capturing high-level semantic relationships in a series of interdependent input values that can be useful for various tasks, such as text completion or predicting the next word in a sentence. Compared to standard independent and identically distributed datasets, elements in the sequence follow a certain order and are not independent of each other. Typical machine learning algorithms and standard feedforward networks fall short in sequence modeling due to their inability to handle the order and keep the memory of past seen samples~\cite{sutskever2014sequence}. Natural language processing, time-series analysis, and  predictive process monitoring are the key areas for sequence modeling.  

\textbf{Recurrent neural networks (RNNs)} introduced an internal memory state to retain the memory of past inputs. The hidden layer within a recurrent network receives input from the input layer and the previous hidden layer's output at each timestep. However, RNNs cannot maintain the context information for long input sequences and suffer from gradient vanishing or exploding problems due to the recursive derivation of gradients during model training. 

\textbf{Long Short Term Memory networks (LSTMs)} are special kinds of RNNs with multiple switch gates to avoid the gradient vanishing problem and to remember long-range input dependencies~\cite{hochreiter1997long}. Even though LSTMs perform better than RNNs, they suffer from similar limitations as RNNs when an input sequence become excessively long. Additionally, LSTMs are computationally expensive to train due to long sequential gradient paths which makes it harder to parallelize them.  

\begin{wrapfigure}{r}{0.4\textwidth}
\centering
\includegraphics[width=0.5\textwidth]{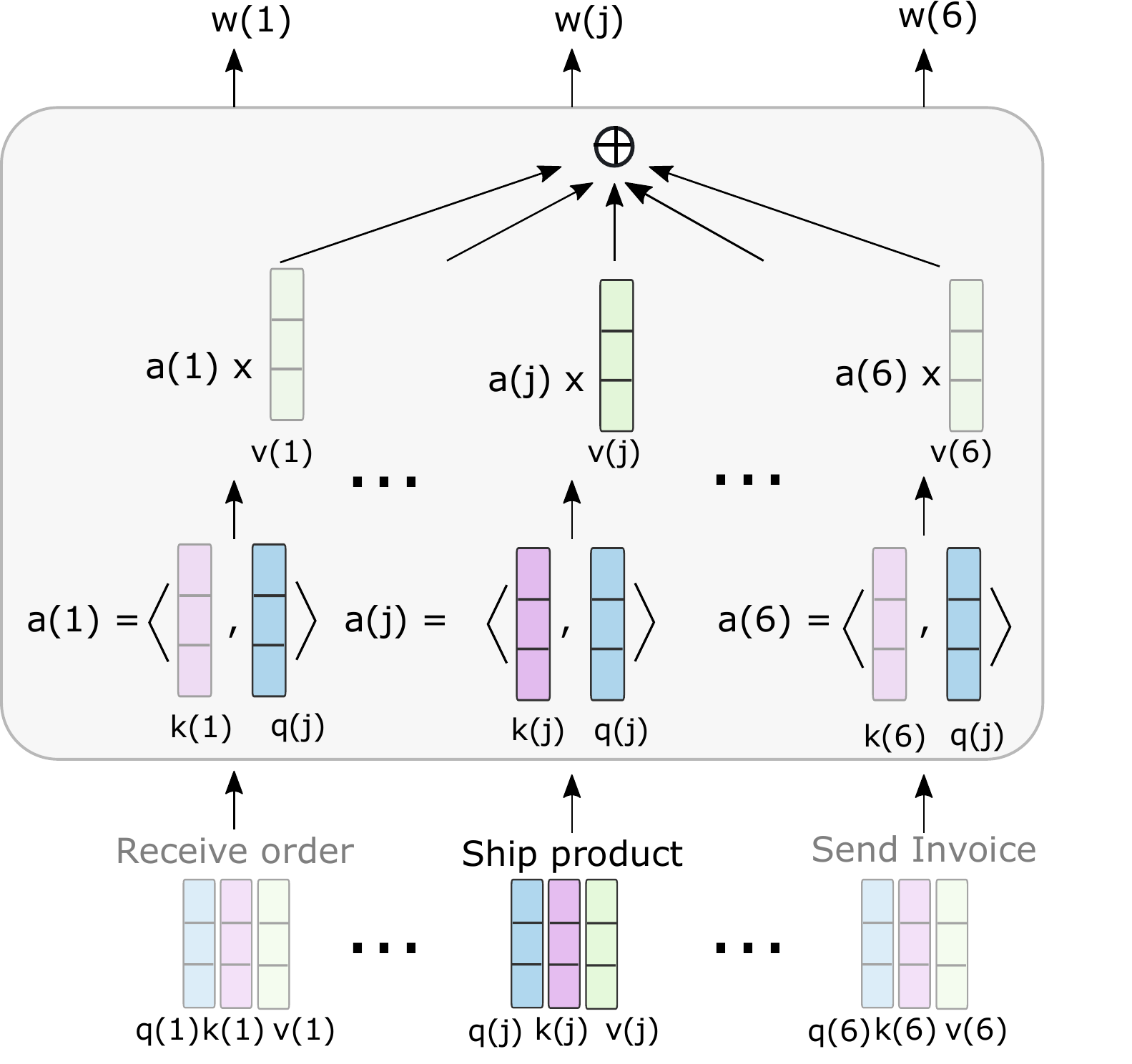}
\caption{\label{fig:attension} An example of Self-attention mechanism to learn attention representation vector of \textit{Ship product} event.}
\end{wrapfigure}

\textbf{Transformer} is introduced by Vaswani et al.~\cite{vaswani2017attention} for neural machine translations. Essentially, the transformer replaced the recursive approach of previously introduced recurrent networks with the \textbf{self-attention mechanism}. It enables the transformer to reason over long-range dependencies and draw generic representations between input and output. Self-attention mechanism decides the importance of all tokens in a sequence with respect to the input token. In Figure~\ref{fig:attension}, we introduce the self-attention mechanism with the help of an example. A model based on self-attention takes a trace having six events as input. Each event is represented with embedding vectors called a query, key, and values. To compute the attention representation vector $w_j$ of~\textit{Ship product} event, we must measure how it relates to the other events in a trace. For this, we take the dot product of the query vector of interest with the keys of all input vectors, resulting in a vector of weights $a_j$ for all the input tokens. The attention vector is then computed by taking the weighted sum of $a_j$ with value vectors. In other words, every output is a weighted sum of every input. The attention representation for each input token can be computed in parallel since their operations are independent, thus eliminating the recurrence. A single event can be related to other events in a sequence in multiple ways, such as semantically, temporally, among others. Therefore, the self-attention is projected multiple times to capture all of these dependencies, forming the \textbf{multi-head self-attention}.  

The transformer is mainly used for language modeling; however, it is a generic network that can be adopted for multiple sequence modeling tasks. In this paper, we propose the transformer model to address the PBPM tasks. Further details of ProcessTransformer are given in Section~\ref{sec:approach}. 

\subsection{Predictive process monitoring}
PBPM has emerged as a promising area of process mining, having a wide array of business applications. 
Initially, the research focus has been on examining the process outcome in terms of its duration and successful completion. Multiple process analytics techniques based on hidden Markov model~\cite{pandey2011test}, finite state machines~\cite{folino2012context}, stochastic Petri nets~\cite{rogge2013prediction}, and annotated transition systems (using diverse abstractions)~\cite{van2011time} have been proposed. Classical machine learning classification methods, such as random forest and support vector machines, are also adapted to predict the successful completion of a process using the history of event sequences and hand-crafted features~\cite{conforti2013supporting,kang2012periodic}. 

With the automated features learning capabilities of deep neural networks, specifically LSTMs, deep learning methods have been adopted to extract useful representations from large-scale temporal sequential event logs to solve various tasks. Successful application of deep sequence modeling has been explored earlier by Evermann et al.~\cite{evermann2016deep} for predicting the next event using a shallow LSTM model and embedding technique for handling categorical variables.
A similar LSTM model architecture with one-hot vector encoding was employed by Tax et al.~\cite{tax2017predictive} to predict the next activity with its associated timestamp, remaining process duration, and process suffix.
Likewise, the same process monitoring problems were addressed by Camargo et al.~\cite{camargo2019learning} using the composition of LSTM to support both categorical and numeric features. Additional temporal features are proposed in~\cite{navarin2017lstm} and~\cite{nguyen2020time} to improve the predictive capabilities of existing deep models. An extension of LSTM with attention mechanism is used in~\cite{wang2019outcome} for the process outcome prediction task.

% Lin et al.,~\cite{lin2019mm} proposed an encoder-decoder framework based on LSTM to solve multiple process monitoring problems in a multi-task setting.

Although LSTM has been a popular choice due to its sequence modeling characteristics, other architectural variants of deep neural networks and analytical techniques are also explored in few studies. Prominently, Khan et al.,~\cite{khan2018memory} introduced memory augmented neural networks as a recommendation tool for tackling complex process analytic problems. Pasquadibisceglie et al.,~\cite{pasquadibisceglie2019using} proposed a data engineering approach to transform events temporal data to spatial image-like structure in order to use the convolution neural networks (CNN). Similarly, Mauro et al.,~\cite{di2019activity} adapted the inception architecture of CNN for sequential data to address the next activity prediction problem. Pauwels et al.~\cite{pauwels2020bayesian} presented a Bayesian technique to predict the next event. Taymouri et al.,~\cite{taymouri2020encoder} adapted generative adversarial nets (GANs) with Gumbel-Softmax distribution to use them for (categorical) suffix generation and remaining time prediction. Bohmer et al.~\cite{bohmer2020logo} proposed combining local and global techniques using sequential prediction rules for the next event prediction. For further relevant work on PBPM, an interested reader may refer to~\cite{rama2020deep} and~\cite{weinzierl2020empirical} survey studies. 

%However,  GANs are infamous for being notoriously unstable during training and they might be inordinate choice as a pre-training strategy when synthesizing data is not a core focus. %Additioanlly, , makingit a great challenge to use them in practice, for now~\cite{thanh2019improving}. Therefore, we don't refer to ~\cite{taymouri2020encoder} for comparisons in this study. 
\section{Preliminaries}
\label{sec:pre}
In this section, we introduce concepts that will be used to define the problem and data preprocessing for predictive process modeling in the subsequent sections. We follow the standard notations provided in~\cite{tax2017predictive} and~\cite{rama2020deep}. 
\vspace{-0.5cm}
\subsubsection*{Definition 1 (Event)}  Let $\mathit{A}$ be the set of activities, $\mathit{C}$ the set of cases, $\mathit{T}$ the time domain and $D_1,.., D_m$  the set of related attributes where $m > 0$. An \texttt{event} is a tuple $e = (a, c, t, d_1, \ldots, d_m)$, where $a\in A$, $c\in C$, $t\in T$ and $d_i \in \{ D_i\}$ with $i~\in [1,m]$. 
\vspace{-0.5cm}
\subsubsection*{Definition 2 (Trace, Events Log)} Let $\pi_A$, $\pi_C$, and $\pi_T$ be functions that map an event $e = (a, c, t, d_1, \ldots, d_m)$ to an activity, as $\pi_A (e) = a$,  to a unique case identifier, as $\pi_C (e) = c$ and to a timestamp, as $\pi_T (e) = t$. A  \texttt{trace} is defined as a finite non-empty sequence of events $\sigma = \langle e_1, e_2, \ldots, e_n \rangle$, such that $\forall~e_i, e_j \in  \sigma$, it must hold that: the events within a trace $\sigma$ must have same case id, i.e. $ \pi_C (e_i) =  \pi_C (e_j)$ and time should be non-decreasing, i.e. $\pi_T (e_j) \geq  \pi_T (e_i)$ for $j > i$. We say that a trace $\sigma = \langle e_1, e_2, \ldots, e_n \rangle$ has length $n$, denoted $|\sigma|$. An \texttt{event log} is collection of traces $L = \{ \sigma_1, \sigma_2, \ldots, \sigma_l \}$. We say that a collection $L = \{ \sigma_1, \sigma_2, \ldots, \sigma_l \}$ has size $l$, denoted $|L|$. 
\vspace{-0.4cm}
\subsubsection*{Definition 3 (Activity  Prediction)}
\label{sec:def3}
Let $\sigma$ be a trace $\langle e_1, e_2, \ldots, e_n \rangle$ and $k \in [1,n-1]$ be a scalar positive number.  The event prefix of length $k$, $hd^k$ can be defined as:  $hd^k (\sigma) = \langle e_1, e_2, \ldots, e_k \rangle$. The activity prefix can be obtained by the application of mapping function $\pi_A$ as  $\pi_A(hd^k(\sigma)) = \langle \pi_A(e_1), \pi_A(e_2), \ldots, \pi_A(e_k) \rangle$. Activity prediction is the definition of a function $\Theta_a$ that takes event prefix $hd^k (\sigma)$ where $k\in [1,n-1]$, and predicts the next activity $e'$, i.e.: \\
\begin{center}
$\Theta_a (hd^k (\sigma)) = \pi_A (e'_{k+1})$    
\end{center}
\vspace{-0.5cm}
\subsubsection*{Definition 4 (Event Time Prediction)}
\label{sec:def3}
Let $\sigma = \langle e_1, e_2, \ldots, e_n \rangle$ be a trace of length $n$. To extract the time-related features of the last event $e_n$ of that trace we define the functions:
\begin{align*}
fv_{t1}(\sigma) &= 
 \begin{cases} 0 & \text{if}~|\sigma| = 1, \\ 
 \pi_T(e_n) - \pi_T(e_{n-1}), & \text{otherwise}.\end{cases} \\
fv_{t2}(\sigma) &=
  \begin{cases} 0 &\text{if}~|\sigma| \in [1,2], \\ 
  \pi_T(e_n) - \pi_T(e_{n-2}), & \text{otherwise}.\end{cases} \\
fv_{t3}(\sigma) &= 
 \begin{cases} 0 & \text{if}~|\sigma| = 1, \\ 
 \pi_T(e_n) - \pi_T(e_0)  \hspace{0.5cm} & \text{otherwise}.\end{cases} \\
\end{align*}
The $fv_{t1}$ feature represents the time difference between the previous event and the current event of a trace. The $fv_{t2}$ feature contains the time difference between current event time and time of an event before the previous event. Finally, $fv_{t3}$ depicts the approximate time passed since the case has initiated. (Note that we say the approximate time, because, due to the fact that we only have the completion time of each event, we do not know the duration of the first event or the time the case waited for the first event to occur.) Event time prediction is the definition of a function $\Theta_t$ that takes event prefix $hd^k (\sigma)$ where $k\in [1,n-1]$, and predicts the time moment at which the next activity will occur, i.e.: 
\begin{center}
$\Theta_t (\sigma', fv_{t1}(\sigma'),fv_{t2}(\sigma'),fv_{t3}(\sigma')) = \pi_T (e'_{k+1})$, where $\sigma' = hd^k (\sigma)$
\end{center}
\vspace{-0.5cm}
\subsubsection*{Definition 5 (Remaining Time Prediction)}
\label{sec:def5}
Let $\sigma = \langle e_1, e_2, \ldots, e_n \rangle$. Remaining time prediction is the definition of a function $\Theta_{rt}$ that takes event prefix $hd^k (\sigma)$ where $k\in [1,n-1]$, and predicts the remaining time of the case, i.e.: 
\begin{center}
$\Theta_{rt} (\sigma', fv_{t1}(\sigma'),fv_{t2}(\sigma'),fv_{t3}(\sigma')) = \pi_T (e_{n}) - \pi_T (e_k)$, where $\sigma' = hd^k (\sigma)$
\end{center}

Note that $fv$ functions are applied manually for feature extraction as a preprocessing step, whereas $\Theta$ functions are learned in an end-to-end manner with~\pt. 
\section{Process Transformer}
\label{sec:approach}
Real-life event logs present temporally sequential data, which is complex, variable, has extensive dependencies due to multiple control flows. Recurrent neural networks, such as LSTM, struggle to reason over long-range sequences due to the limited size of a context vector, as noted in~\cite{paperno2016lambada}. This paper addresses the problem of PBPM to predict the next activity, event time, and remaining time of a process under execution, i.e., using deep learning to learn the functions $\Theta_a$, $\Theta_t$, and $\Theta_{rt}$ as they are defined in Definitions~\hyperref[sec:def3]{3-5}. To that end we propose the \pt. Notably, while several deep learning-based process monitoring methods~\cite{tax2017predictive,camargo2019learning,pasquadibisceglie2019using} exist, which learn a predictive model based on varying prefixes length of event sequences, we develop a deep neural network that considers possible prefixes altogether for training and inference. 

% We propose a transformer model, named~\pt, based on the self-attention mechanism introduced by~\cite{vaswani2017attention}. 

% \subsubsection*{Problem description}
% Here, we address the problems of PBPM to predict the next activity, event time, and remaining time of a process under execution. 
% The datasets are prepared for each problem using the functions defined in Definitions~\hyperref[sec:def3]{3-5}. 
% We also manually extract three temporal features to introduce a deep neural network about the typical duration of events in a sequence. 
% The proposed~\pt model aims to extract high-level representations from raw large-scale event logs data with minimal prepossessing.  Notably, while several deep learning-based process monitoring methods~\cite{tax2017predictive,camargo2019learning,pasquadibisceglie2019using} learn a predictive model based on varying prefixes length of event sequences, we develop a single deep neural network that considers possible prefixes altogether for training and inference.   

\begin{figure}[!b]
\centering
\includegraphics[width=0.8\textwidth]{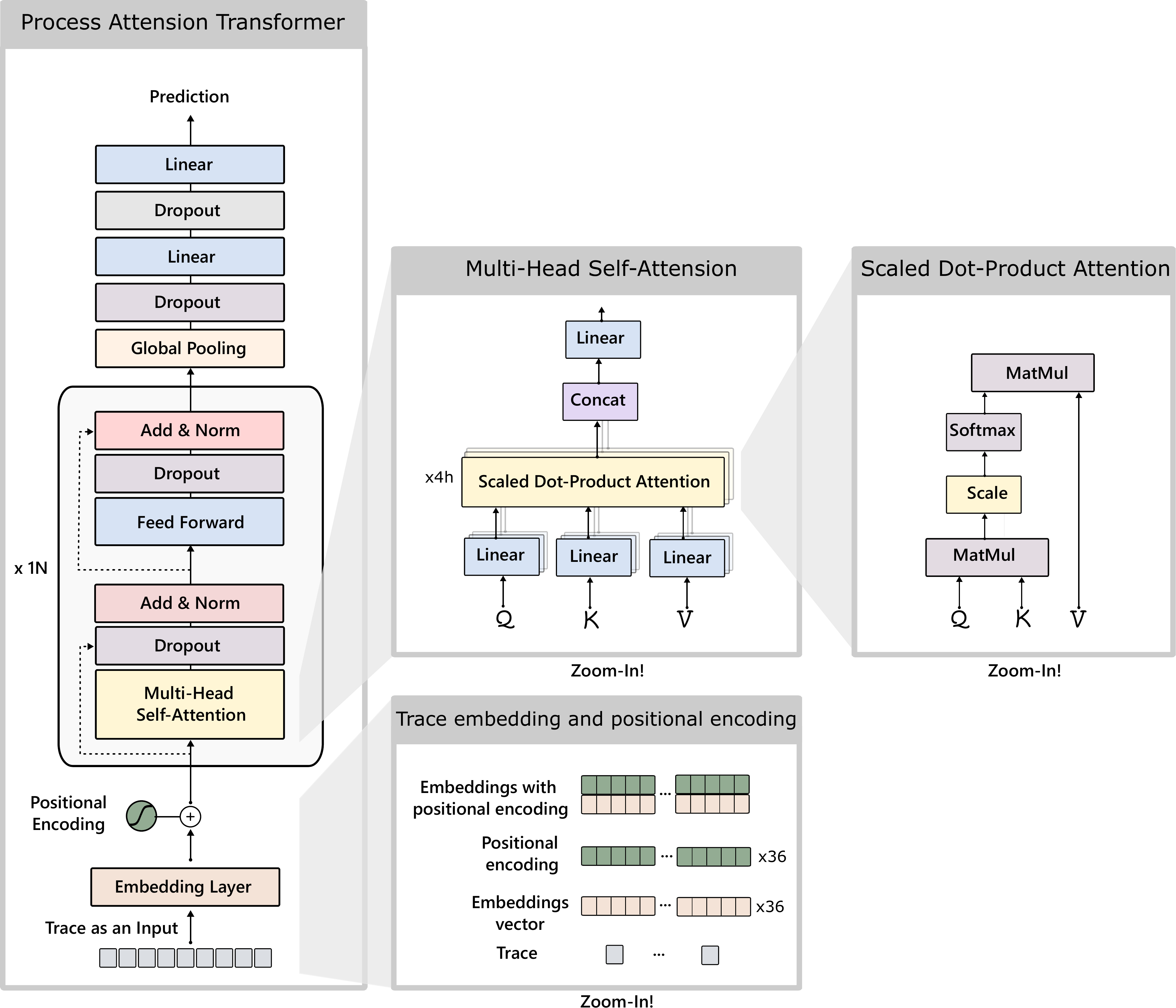}
\caption{Model Architecture of Process Transformer. }
\label{fig:model}
\end{figure}

%\subsection{Process Attention Transformer}
% Real-life event logs present temporally sequential data, which is complex, variable, has extensive dependencies due to multiple control flows.
% Recurrent neural networks, such as LSTM, struggle to reason over long-range sequences due to the limited size of a context vector, as noted in~\cite{paperno2016lambada}. 
% We propose a transformer model, named~\pt, based on the self-attention mechanism introduced by~\cite{vaswani2017attention}. 

Figure~\ref{fig:model} provides a high-level overview of the model architecture. 
The~\pt has $\mathcal{N}$~\textit{attention blocks}, which take positional encoded input sequence and pass learned representation to a pooling operation and then to a fully connected layer. We use single attention block for~\pt. The attention block comprises~\textit{multi-headed attention} layers, followed by feed-forward layers having residual connections, dropout, and a normalization layer. 
In the following, we explain the most important building blocks of our model. \\

\textbf{Trace Embedding and Positional Encoding:} Starting from a trace, the network learns vector embedding for each event as shown in the respective block in Figure~\ref{fig:model}. 
It essentially maps the categorical input to the vector of continuous representations.
The network learns vector embedding and projects them into a transformed space where similar events in a semantic sense are mapped closer to each other. 
The benefit of using embedding in this way is that it eliminates the problem of high-dimensionality encountered in a one-hot encoding scheme. For example, while a one-hot encoding would require a binary vector for each event depending on the size of vocabulary (unique event instances), the embedding layer learns representation of semantically similar events by mapping them individually to a vector space.  %$36$ dimensional vector.% a more efficient encoding in which ... 

It is worth noting that the~\pt does not use recurrence as former approaches (e.g., in~\cite{tax2017predictive,camargo2019learning}).
This property enables efficient training but at the cost of omitted positional information of events from traces. 
The default Transformer architecture proposed to add \textit{positional encoding} along with an input embedding to inform the model about the relative positioning of each token in a sequence.
We choose the \textit{36}-dimensional vector encoding to represent the relative positioning of an event in a trace. 
The input embedding and position encoding have the same dimension to allow for their summation at the next step. 
The neural model learns to attend to  positional encoding along with input embedding in an end-to-end manner during learning to solve a specific task. \\

\begin{figure}[!b]
\centering
\includegraphics[width=0.8\textwidth]{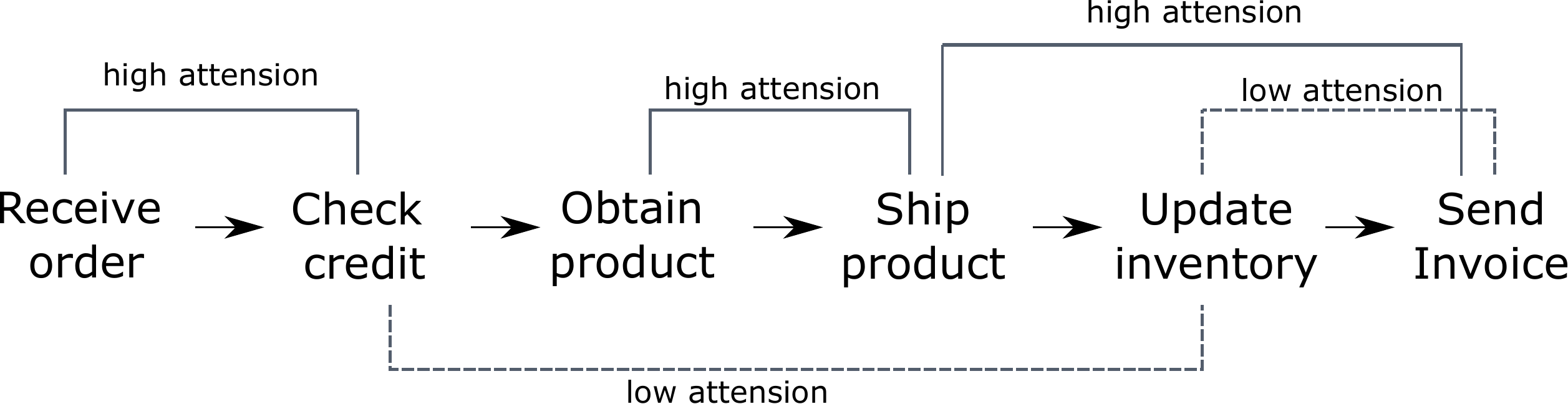}
\caption{Illustration of self-attention mechanism on a trace. The model learns attention scores of events for solving a particular predictive monitoring task. }
\label{fig:example}
\end{figure}

\textbf{Self-Attention (Scaled Dot-Product)}
%The attention mechanism is inspired by the human cognitive attention system.
%For instance, our visual system focuses on specific parts of scenery while ignoring the less relevant information.
%Similarly, 
Self-attention models learn to selectively attend to only important parts of a trace to compute the robust representation. It enables the Transformer to reason over long-range dependencies and draw generic representations between input and output. We illustrate the self-attention mechanism with the help of an example trace of \textit{order management process} in Figure~\ref{fig:example}. It shows that the events such as \textit{receive order} and \textit{check credit}, \textit{obtain product} and \textit{ship product} are related and will have high relative attention scores with respect to each other.  The events like \textit{update inventory}, \textit{send invoice}, and \textit{check credit} are semantically different in the context of order management and may have low attention scores. This self-attention enables the model to pay attention to important event(s) of a trace in order to better solve the predictive monitoring tasks.    

 The attention mechanism initially creates three vectors called  query $q$, key $k$ and value $v$ for each input embedding i.e. $x_1$, as shown in Figure~\ref{fig:model}.
The neural model learns the representations for these vectors. %by multiplying $x_1$ to three respective weight matrices, i.e., $W^k$, $W^v$, and $W^v$.
%The next step is to obtain the \textit{attention} score of each input embedding with respect to an overall input sequence.
A~\textit{self-attention} function maps a query $q$ to a set of key-value pairs, denoted as $k$ and $v$, to obtain a weighted sum of values called output, denoted by $\mathcal{Z}$.
%The most common mapping functions to compute $\mathcal{Z}$ are additive attention and dot-product (multiplicative) attention. 
Following~\cite{vaswani2017attention}, we adopt scaled dot-product attention to compute $\mathcal{Z}$ as follows:
\begin{equation*}
    \text{Attention}(\textit{Q, K, V}) =\mathcal{Z} =  \text{softmax}(\frac{QK^T}{\sqrt{d_k}})V  
\end{equation*} 
where $Q$, $K$, and $V$ are matrices of $q$, $k$, and $v$ vectors packed into respective matrices for efficient computations.
The attention scores are scaled by dividing to $\sqrt{d_k}$ i.e., dimension of key $k$, for stable gradient computations.
Afterward, the~\textit{softmax} function is applied to normalize the scores and obtain the importance of each token.
Finally, the softmax score is multiplied with the value $V$ matrix to give the model capability to focus on which words to \textit{attend} to in a sequence and eliminate less relevant tokens. \\
%  (due to multiplication to smaller softmax values)

\textbf{Multi-Head Self Attention}
A single event can be related to other events in a trace in multiple ways, such as semantically, temporally, among others. 
In order to introduce a model with different representation subspaces at different positions~\cite{vaswani2017attention}, the query, key, and values are linearly projected $h$ times as shown in Figure~\ref{fig:model}.
The scaled dot-product attention is performed in parallel for each of the projection, forming the~\textit{multi-headed attention} as follows:
\begin{align*} 
    \text{MultiHead}(Q, K, V) = \text{Concat}(head_1, \ldots, head_h)W^O \\
    head_i = \text{Attention}(QW_i^Q, KW_i^K, VW_i^V )
\end{align*}

\noindent where the output of each $head_i$ is linearly concatenated and multiplied by weight matrix $W^O$, that is learned by the deep neural network. 

A dropout layer follows the output of the multi-head self-attention block to limit over-fitting, and a layer normalization is applied across the layer's features. %~\cite{ba2016layer}
Afterward, a position-wise feed-forward layer followed by a dropout and layer normalization is applied. 
The attention blocks $\mathcal{N}$ make use of the residual connection, depicted by dashed lines in Figure~\ref{fig:model}, to enable gradients to skip non-linear functions, thus avoiding the vanishing gradient problem.% as shown in ResNet architecture.%~\cite{he2015deep}.
~The rest of the layers after the attention block $\mathcal{N}$ are mainly network design choices, which are covered in detail in the following section. %with different activation function and parameters. The deails are provided in the following section.  
%where the parameter metrics are $W^Q_i, W^K_i \in \mathbb{R}^{d_{model} \times d_k}$,
%W^V \in \mathbb{R}^{d_{model} \times d_v}$, 
%$W^O\in \mathbb{R}^{hd_v \times d_{model}}$.

% sine and cosine encoding function, which is represented as follows:
% \begin{align*} 
%  \text{PE}_{(pos, 2_i)} = \sin (pos /10000^{2i/d_{model}}) \\
%  \text{PE}_{(pos, 2_i+1)} = \cos (pos /10000^{2i/d_{model}})
% \end{align*}
% where $pos$ is the position and $i$ is the dimension. These functions are applied alternatively to create positional vectors from input embedding. The sinusoidal functions have linear properties that enable the model to learn to attend the relative positioning. 
\vspace{-0.3cm}
\subsubsection{Network Design and Implementation}  
Given an event log, we create a word dictionary to encode the event names numerically.
The traces that are shorter than $L_{|\sigma_n|}$ (i.e., the maximum length of a trace in event logs) are padded with zero to have a fixed-length input. 
We transform each event in a trace using a learnable \textit{distributed embedding} of $36$ units and a~\textit{positional encoding} with the same dimension. 
The embedding outputs are then fed to a~\textit{multi-headed attention} block with $h = 4$ ($h$ being the number of heads) in order to learn general-purpose representations at different input positions. 
We apply a global max-pooling on the last layer of the attention block to aggregate features, followed by a dropout with a rate of $0.1$.
We also use dense layers with $32$ and $128$ hidden units having ReLU activation consecutively.
A fully-connected layer with hidden units corresponding to the output dimension of a predictive task is used, with either softmax or linear activation functions (depending on the target task), to produce output.

We employ categorical cross-entropy loss for the next activity prediction task. Likewise, we use the log-cosh loss function for regression-based tasks, i.e., for the event time and remaining time prediction problem. Here, we concatenate the output of the attention block with scaled numeric temporal attributes. The rest of the architecture remained the same across tasks. 

% The model learn embeddings of the input sequence using the embedding layer. Since there is no recurrence in the transformer, we provide the position encoding by using some sine, cosine function for even and odd input sequence. The core of the Transforer is the attention block represented by N which can be more than one. The attention block consist of self-attention mechanism which enables the model to learn to attend to specific word in an input sequence. In contrast to typical sequence modeling architecture like GRU, RNN and LSTSM, the attention-mechanism has indefinite memory which enables the network to attend to large/length input temporal sequences. The self-attention mechism also enables the learning of global general-purpose representations. 

\section{Evaluation}
\label{sec:evaluation}
We evaluate the efficacy of the proposed~\pt on nine real-life event logs.
We also provide comparisons to established benchmarks reported in~\cite{rama2020deep}.
This section introduces the experimental setup, including datasets and evaluation metrics, followed by input preprocessing and network design details.
We conclude the section by providing evaluation results of three predictive process monitoring tasks namely next activity, next event time and remaining time prediction of a running case.

%  and results. Finally, we report results in the last subsection.  The experiments were performed using two Nvidia Tesla T4 GPUs, and 64GB RAM on the Google Cloud Platform. 

\subsection{Experimental setup}
\subsubsection*{Datasets} The experiments were conducted using event logs publicly available at the 4TU Research Data repository\footnote{\url{https://data.4tu.nl/categories/_/13500?categories=13503}}.
Some of these datasets have been widely used to evaluate process  monitoring tasks (e.g.~\cite{camargo2019learning,lin2019mm,tax2017predictive,evermann2016deep}). Table~\ref{tab:datasets} provides the descriptive statistics of considered event logs. %A brief description of each is also provided below.

\begin{table}[!htbp]
\caption{Descriptive statistics of event logs used for evaluations. Time-related characteristics are reported in days.}
\label{tab:datasets}
\centering
\resizebox{0.8\textwidth}{!}{%
\begin{tabular}{lccccccc}
\toprule
Datasets & \multicolumn{1}{l}{Cases} & \multicolumn{1}{l}{Events} & \multicolumn{1}{l}{Activities} & \begin{tabular}[c]{@{}c@{}}Max case \\ length\end{tabular} & \begin{tabular}[c]{@{}c@{}}Avg. case\\  length\end{tabular} & \begin{tabular}[c]{@{}c@{}}Max case \\ duration \end{tabular} & \begin{tabular}[c]{@{}c@{}}Avg. case\\  duration \end{tabular} \\
\cmidrule{1-8}
\textit{Helpdesk}~\cite{polato_2017} & 4,580 & 21,348 & 14 & 15 & 4.66 & 60 & 40.69 \\
\textit{BPIC12}~\cite{dongen_2012} & 13,087 & 262,200 & 24 & 175 & 20.03 & 13 & 8.01 \\
\textit{BPIC12w}~\cite{dongen_2012} & 9,658 & 170,107 & 7 & 156 & 17.61 & 132 & 10.5 \\
\textit{BPIC12cw}~\cite{dongen_2012} & 9,658 & 72,413 & 6 & 74 & 7.497 & 82 & 10.46 \\
\textit{BPIC13}~\cite{BPIC132014} & 7,554 & 65,533 & 13 & 123 & 8.6754  & 768 &  11.948\\
\textit{BPIC20d}~\cite{Dongen2020d} & 10,500 & 56,437 & 17 & 24 & 5.37 & 47 & 11.16 \\
\textit{BPIC20i}~\cite{Dongen2020i} & 6,449 & 72,151 & 34 & 27 & 11,187 & 737 & 84.15 \\
\textit{Hospital}~\cite{mannhardt_2017} & 100,000 & 451,359 & 18 & 217 & 4.51 & 1034 & 127.24 \\
\textit{Traffic fines}~\cite{fines15} & 150,370 & 561,470 & 11 & 20 & 3.73 & 4373 & 342.67 \\
\bottomrule
\end{tabular}%
}
\end{table}
\vspace{-0.4cm}
\subsubsection*{Evaluation metrics}  For the next activity prediction task, \textit{accuracy} is a commonly used metric as reported in several studies~\cite{tax2017predictive,khan2018memory,pasquadibisceglie2019using,evermann2016deep}. \textit{Accuracy} is essentially computed by taking a fraction of correctly predicted samples to the total number of samples. However, in the case of an imbalanced dataset, \textit{accuracy} as performance metric can be misleading. Therefore, we report weighted accuracy and weighted \textit{F-score} in the paper. The \textit{weighted} aspect considers the data imbalance of the target class and assigns the weights to data samples accordingly. 
The \textit{F-score} is a combination of precision (or positive predictive value)
and recall (sensitivity) measures~\cite{sokolova2009systematic}. 
The precision determines the exactness of the model, whereas the recall provides a measure of the model's completeness. 
\textit{F-score} is calculated as follows: 
\[ \text{F-score} = 2 \times \frac{\text{precision} \times \text{recall}}{\text{precision} + \text{recall}}\]

For next event time and remaining time prediction having continuous target output, we compute \textit{mean absolute error (MAE)} as follows: %which represents the mean of the absolute error of individual predictions on overall test set. 

\[ \text{MAE} = \frac{\sum_{i=0}^{n} |y_i - \hat{y_i}|}{n} \]

\noindent where $y_i$ is the predicted value from the model, $\hat{y_i}$ is the desired output, and $n$ is the total number of samples in the test set.

\subsection{Data preprocessing and training setup}  
The event logs are first chronologically ordered to simulate the reality in which a model uses past traces to monitor the future performance of running traces.
Each dataset is split into $80$\% and $20$\% for training and testing sets, respectively, while preserving its temporal order.
Additionally, $20$\% of the data from the training split is used for validation and hyper-parameter tuning during the learning phase.    

For training and evaluation of~\pt network, following key points are important: \begin{itemize}
    \item  We use the raw event logs data with minimal preprocessing. This means we did not selectively filter out events with specific k-prefixes, i.e. $hd^k (\sigma)$, unlike noted in ~\cite{tax2017predictive,taymouri2020encoder}. The use of all prefixes enables the model to perform process monitoring tasks for extremely small (e.g., single event) to very lengthy running cases.
    \item The model is trained for $100$ epochs with an ADAM optimizer~\cite{kingma2014adam} and a learning rate of $10^{-2}$ for all the considered tasks. Furthermore, we explore the impact of the batch size and the number of attention heads. We report the results in the subsequent section with the optimal parameters configuration found on the validation set. 
    \item We evaluate the model's performance on the test set iteratively for each k-prefixes, i.e., $hd^k (\sigma)$. This is to illustrate the model's predictive capability given the limited size of prefix, e.g. only single event, as an input. We iteratively compute the performance metric score, such as  accuracy, MAE, for each k-prefixes and reports the average results across all prefixes in the following  section.   %depending on the predictive task. The average  ofThe trained model takes the samples from testset as input having We denote this for a single case $\sigma$ as follows:
    % \vspace{-0.5cm}
    % \begin{center}
    %  \[ \frac{1}{|\sigma|} \sum_{k=1}^{|\sigma|} \Theta (hd^k(\sigma)) \]
    % \end{center}
    % where $K$ is length of a case, $\Theta$ is a predictive model that takes a prefix as input and produces a probability distribution over activities in case of a next activity prediction task. The output of the model is averaged across all the k-prefixes of a case. In the case of the next event and remaining time prediction tasks, the models take temporal features as input in addition to the prefix.    
\end{itemize}

\subsection {Results}
We report the experimental results to assess the performance of~\pt for prediction of the next activity, event time, and remaining time of a running case.
Table~\ref{tab:all} reports accuracy, F-score, and MAE for nine datasets on three process monitoring tasks.
Due to large event logs and time constraints, we do not provide baseline comparison scores for four datasets, namely, BPIC20i, BPIC20d, hospital, and traffic fine logs.
For the other five datasets, we report performance comparisons with other related studies. We adopt the baselines scores from the benchmark survey on PBPM as reported in~\cite{rama2020deep}.

\begin{table}[!htbp]
\renewcommand{\arraystretch}{1.2}
\caption{Performance evaluation scores of \pt for nine event logs for three predictive monitoring tasks.}
\label{tab:all}
\centering
\resizebox{0.6\textwidth}{!}{%
\begin{tabular}{lcccc}
\toprule
 & \multicolumn{2}{c}{\textbf{\begin{tabular}[c]{@{}c@{}}Next \\ Activity\end{tabular}}} & \textbf{\begin{tabular}[c]{@{}c@{}}Next \\ Event Time\end{tabular}} & \textbf{\begin{tabular}[c]{@{}c@{}}Remaining\\ Time\end{tabular}} \\ [0.7ex] 
 \cline{2-5} 
 & \multicolumn{1}{l}{\textbf{Accuracy}} & \multicolumn{1}{l}{\textbf{F-score}} & \textbf{MAE} & \textbf{MAE} \\ [0.7ex]
\midrule
Helpdesk & 85.63 & 0.82 & 2.98 & 3.72 \\
BPIC12 & 85.20 & 0.83 & 0.25 & 4.60 \\
BPIC12w & 91.51 & 0.91 & 0.37 & 4.87 \\
BPIC12cw & 78.48 & 0.77 & 0.82 & 5.14 \\
BPIC13i & 62.11 & 0.60 & 0.99 & 8.36 \\
BPIC20d & 86.07 & 0.84 & 1.22 & 2.44 \\
BPIC20i & 93.35 & 0.91 & 3.26 & 10.68 \\
Hospital & 85.83 & 0.82 & 9.33 & 44.87 \\
Traffic   fines & 90.00 & 0.87& 40.28 & 98.24 \\
\bottomrule
\end{tabular}%
}
\end{table}
 % We do not re-run the experiments of previous studies, since most of the studies has focus on introducing encoding techniques and temporal features to improve performance instead of methodolical advancements.  

\subsubsection*{Next Activity Prediction}
Table~\ref{tab:NA} reports the (weighted) accuracy for the next activity prediction task on five real-life event logs. 
Except for BPIC13, ~\pt consistently outperforms multiple baselines reported in the literature for the next activity task. Notably, we achieve $86$\%, $85$\%, $91$\% and $78$\% accuracy scores for \textit{Helpdesk}, \textit{BPI2012}, \textit{BPI2012w}, and \textit{BPI2012cw} datasets, respectively. The low accuracy of BPIC13 can be attributed to the fewer but lengthy cases in the logs.  We also show that on average across all the datasets,~\pt outperforms other approaches. 

\begin{table}[!htbp]
\caption{Accuracy score (in \%) and averaged scores across all datasets for next activity prediction task (Higher is better). The baseline scores for comparison are taken from~\cite{rama2020deep}.}
\label{tab:NA}
\centering
\renewcommand{\arraystretch}{1.2}
\resizebox{0.9\textwidth}{!}{%
\begin{tabular}{lcccccc}
\toprule
 & \textbf{Helpdesk} & \textbf{BPIC12} & \textbf{BPIC12w} & \textbf{BPIC12cw} & \textbf{BPIC13} & \multicolumn{1}{c}{\textbf{Avg.}}   \\ 
\midrule
Tax et al.~\cite{tax2017predictive} & 75.06 & 85.20 & 84.90 & 67.80 & 67.50 & 76.09 \\
Khan et al.~\cite{khan2018memory} & 69.13 & 82.93 & 86.69 & 75.91 & 64.34 & 75.80 \\
Camargo et al.~\cite{camargo2019learning} & 76.51 & 83.41 & 83.29 & 65.19 & 68.01 & 75.28 \\
Evermann et al.~\cite{evermann2016deep} & 70.07 & 60.38 & 75.22 & 65.38 & 68.15 & 67.84 \\
Mauro et al.~\cite{di2019activity} & 74.77 & 84.56 & 85.11 & 65.01 & \textbf{71.09} & 76.11 \\
Pasquadibisceglie et al.~\cite{pasquadibisceglie2019using} & 65.84 & 82.59 & 81.59 & 66.14 & 31.10 & 65.45 \\ [0.7ex]
\midrule
\textbf{\pt} & \textbf{85.63} & \textbf{85.20} & \textbf{91.51} & \textbf{78.48} & 62.11 & \textbf{80.58} \\
\bottomrule
\end{tabular}%
}
\end{table}

\noindent It is worth noting that we achieve better generalization and performance without performing extensive preprocessing in terms of removing incomplete process traces and $1$-sized prefix having a single event only.
Importantly, the model utilizes only event prefixes as an input without utilizing any hand-crafted features.
These design choices are made to report the realistic analysis and  illustrate the powerful learning capabilities of~\pt even with a single event prefix, duplicate activities, and excessively long process traces.
Given this, a direct comparison with some of the proposed methods from literature is unjust due to inconsistent data preprocessing and additional input features for learning, which can be missing in event logs in a real-world setting.
For instance, ~\cite{di2019activity} utilize additional event attributes, such as timestamp for event label prediction.
Similarly,  ~\cite{tax2017predictive,pasquadibisceglie2019using,khan2018memory} performed excessive preprocessing on the \textit{Helpdesk} and \textit{BPI2012} datasets in terms of eliminating process traces depending on their number of events and their duration. It results in a predictive model trained and evaluated on an ideal dataset, which does not reflect real-life complex event logs.  Furthermore, we do not provide a comparison against~\cite{taymouri2020predictive}, as it utilizes additional synthetic data for model training. However, our technique is complementary and can be combined with GANs to improve performance further.

 \vspace{-0.4cm}
\subsubsection*{Event Time Prediction} In Table~\ref{tab:NT}, we present the MAE in days for the event time prediction task against multiple baselines. Our approach has achieved the lowest MAE on average for all considered datasets compared to the previously proposed methods. Besides algorithmic specifications, our approach is different to~\cite{tax2017predictive,nguyen2020time,khan2018memory,bohmer2020logo} in following aspects. We deal with the event time prediction as an independent task as opposed to the common multi-task approach. This is because the multi-task approach requires dual loss optimization, and there is no guarantee that it will perform better than its single-task counterpart~\cite{zhang2017survey}. We also create additional temporal features (see Definition~\hyperref[sec:def4]{4} in Section~\ref{sec:approach})  to equip model with a sense of events' duration. To summaries,~\pt obtains MAE of $2.98$, $0.25$, $0.37$, $0.82$ and $0.99$ for \textit{Helpdesk}, \textit{BPI2012}, \textit{BPI2012w}, \textit{BPI2012cw} and \textit{BPIC13} datasets, respectively.

\begin{table}[!htbp]
\caption{MAE (in days) and averaged scores across all datasets for event time prediction task (Lower is better). The baseline scores for comparison are taken from~\cite{rama2020deep}.}
\label{tab:NT}
\centering
\renewcommand{\arraystretch}{1.1}
\resizebox{0.8\textwidth}{!}{%
\begin{tabular}{lcccccl}
\toprule
 & \textbf{Helpdesk} & \textbf{BPIC12} & \textbf{BPIC12w} & \textbf{BPIC12cw} & \textbf{BPIC13} & \multicolumn{1}{c}{\textbf{Avg.}} \\ \midrule
Tax et al.~\cite{tax2017predictive} & \textbf{5.77} & \textbf{0.31} & \textbf{0.50} & \textbf{1.20} & \textbf{0.47} & 1.65 \\
Khan et al.~\cite{khan2018memory}& 6.33 & 0.31 & 0.50 & 1.32 & 0.55 & 1.80 \\ \midrule
\pt & \textbf{2.98} & \textbf{0.25} & \textbf{0.37} & \textbf{0.82} & 0.99 & \textbf{1.08} \\
\bottomrule
\end{tabular}%
}
\end{table}

\vspace{-0.4cm}
\subsubsection*{Remaining Time Prediction}
We use the same model architecture and temporal features for the remaining time prediction problem as for event time prediction. Table~\ref{tab:RMT} reports the MAE scores averaged across all the prefixes. Our approach outperforms previous methods on average across considered datasets (see the last column of Table~\ref{tab:RMT}).  Specifically, we obtain MAE of $3.72$, $4.60$, $4.87$, $5.14$, and $8.36$ for \textit{Helpdesk}, \textit{BPI2012}  \textit{BPI2012w}, and \textit{BPI2012cw} and \textit{BPIC3} datasets, respectively.  We note that, the performance difference between Tax et al~\cite{tax2017predictive} and Navarin et al~\cite{navarin2017lstm} is due to difference in encoding techniques and temporal features used, as they both use LSTM as a base model for learning the predictive task.
\begin{table}[!htbp]
\caption{MAE (in days)  and averaged scores across all datasets for remaining time prediction task (Lower is better). The baseline scores for comparison are taken from~\cite{rama2020deep}.}
\label{tab:RMT}
\renewcommand{\arraystretch}{1.1}
\centering
\resizebox{0.9\textwidth}{!}{%
\begin{tabular}{lcccccc}
\toprule
 & \textbf{Helpdesk} & \textbf{BPIC12} & \textbf{BPIC12w} & \textbf{BPIC12cw} & \textbf{BPIC13} & \textbf{Avg.} \\
 \midrule
Tax et al.~\cite{tax2017predictive}& 71.50 & 330.61 & 387.81 & 210.16 & 38.41 & 207.70 \\
Camargo et al.~\cite{camargo2019learning}& 11.15 & 30.56 & 32.03 & 7.97 & 260.64 & 68.47 \\
Navarin et al.~\cite{navarin2017lstm} & 10.38 & 6.13 & 6.63 & 6.48 & \textbf{2.97} & 6.52 \\ \midrule
\pt & \textbf{3.72} & \textbf{4.60} & \textbf{4.87} & \textbf{5.14} & 8.36 & \textbf{5.33}\\
\bottomrule
\end{tabular}%
}
\end{table}

\vspace{-0.7cm}
\section{Conclusion}
\label{sec:conclusion}
The main contribution of our study is a~\pt approach for learning high-level representations directly from sequential event logs data with minimal preprocessing.  We evaluate our approach for the next activity, event time, and remaining time prediction tasks on nine real-life event logs. We show that~\pt can capture long-range dependencies without the explicit need of recurrence as LSTM-based models.
Our approach outperforms several existing baselines in experimental evaluations. Specifically, we achieve an average of above 80\% accuracy on considered datasets for the next activity prediction task while solely using the activity prefix as an input for the model.
Similarly, our approach obtains an average MAE of $1.08$ and $5.33$ for predicting the next event time and completion time of a running case.
Notably, we use minimal data preprocessing and features as an input to illustrate the learning capability of \pt. This is also to emphasis that the \pt can provide optimal predictive performance for real event logs even when additional attributes, such as resources, roles, and others, are missing in the dataset. 

The future work seeks to study how the learned representations can be used for other tasks of interest, including similar trace retrieval, activity recommendations, and process outcome prediction. Another future avenue is to evaluate proposed~\pt with event logs having not only prolonged but largely unique process activity space.  

%Our primary goal was to introduce an attention-based mechanism for representations learning from raw event logs.  %Finally,  there is a need to further improve the temporal aspects of process explore advance avenues of representations learning for process informatics specifically for further improvement of temporal process aspects. \pt has not outperformed baselines for the \textit{BPIC13} dataset which is one of the smallest with lengthy process traces, among the considered datasets. Further experiments are needed to explore if \pt performs well with large event logs only.

\textbf{Reproducibility} 
The source code and supplementary material is publicly available at~\url{https://github.com/Zaharah/processtransformer}.

% \section*{Acknowledgements} 
% \vspace{-0.2cm}
% We thank Google cloud research credits program for providing access to computing resources. 


\bibliographystyle{splncs04}
\bibliography{bibliography}

\end{document}